\documentclass[3p,times,procedia]{elsarticle}

\usepackage{ecrc}

\volume{00}

\firstpage{1}

\journalname{Neural Networks}

\runauth{}

\jid{nc}

\jnltitlelogo{Neural \\ Networks}

\CopyrightLine{2020}{Published by Elsevier Ltd.}

\usepackage[T1]{fontenc}
\usepackage{lmodern}
\usepackage{textcomp}
\usepackage{latexsym}
\usepackage{comment}
\usepackage{amsmath}
\usepackage{amssymb}
\usepackage{amsfonts}
\usepackage{mathrsfs}
\usepackage{amsthm}
\usepackage{listings}
\usepackage{graphicx}
\usepackage{subcaption}
\usepackage{algorithm}
\usepackage{algorithmic}

\makeatletter
\newcommand{\figcaption}[1]{\def\@captype{figure}\caption{#1}}
\newcommand{\tblcaption}[1]{\def\@captype{table}\caption{#1}}
\makeatother

\newtheorem{thm}{Theorem}[section]
\newtheorem{defi}{Definition}[section]

\newtheorem{example}{Example}[section]

\usepackage[figuresright]{rotating}

\begin{document}

\begin{frontmatter}



\dochead{}

\title{The Exact Asymptotic Form of Bayesian Generalization Error in Latent Dirichlet Allocation}


\author[n,t]{Naoki HAYASHI}

\address[n]{Simulation \& Mining Division\\
 NTT DATA Mathematical Systems Inc. \\
1F Shinanomachi Rengakan, 35, Shinanomachi, Shinjuku-ku, Tokyo, 160--0016, Japan\\
hayashi@msi.co.jp}

\address[t]{Department of Mathematical and Computing Science\\
 Tokyo Institute of Technology \\
Mail-Box W8--42, 2--12--1, Oookayama, Meguro-ku, Tokyo, 152--8552, Japan\\
hayashi.n.ag@m.titech.ac.jp}

\begin{abstract}
Latent Dirichlet allocation (LDA) obtains essential information from data by using Bayesian inference.  It is applied to knowledge discovery via dimension reducing and clustering in many fields. However, its generalization error had not been yet clarified since it is a singular statistical model where there is no one-to-one mapping from parameters to probability distributions. In this paper, we give the exact asymptotic form of its generalization error and marginal likelihood, by theoretical analysis of its learning coefficient using algebraic geometry. The theoretical result shows that the Bayesian generalization error in LDA is expressed in terms of that in matrix factorization and a penalty from the simplex restriction of LDA's parameter region. A numerical experiment is consistent to the theoretical result.
\end{abstract}


\end{frontmatter}


\section{Introduction}
Latent Dirichlet allocation (LDA) \cite{David2003LDA} is one of topic models \cite{Gildea1999topic} which is a ubiquitous statistical model used in many research areas. Text mining \cite{David2003LDA, Griffiths2004LDAGS}, computer vision \cite{Li2005LDA4CV}, marketing research \cite{Tirunillai2014LDA4MR}, and geology \cite{Yoshida2018LDA4Geo} are such examples. LDA had been originally proposed for natural language processing and it can extract essential information from documents by defining the topics of the words.
The topics are formulated as one-hot vectors subject to categorical distributions which depend on each document. The parameters of those categorical distributions express the topic proportion and they are the object of inference. In addition, the words are also formulated as one-hot vectors generated by other categorical distributions whose parameters represent appearance probability of the words in each document. This appearance probability is also estimated.

In the standard inference algorithms such as Gibbs sampling \cite{Griffiths2004LDAGS} and variational Bayesian method \cite{David2003LDA}, LDA requires setting the number of topics (the dimension of the topic one-hot vector) in advance. The optimal number of topics in the ground truth is unknown, thus researchers and practitioners face to typical model selection problem; the chosen number of topics may be larger than the optimal one. In this situation, the estimated parameter cannot be uniquely determined. From the theoretical point of view, LDA is non-identifiable, i.e. a map from a parameter set to a probability density function set is not injective. Besides, LDA has a degenerated Fisher information matrix and its likelihood and posterior distribution cannot be approximated by any normal distribution. Such models are called singular statistical models and LDA is one of them.

If a map from parameters to probability density functions is injection in a statistical model, then the model is called a regular statistical model. In a regular statistical model, its expected generalization error is asymptotically equal to $d/2n + o(1/n)$, where $d$ is the dimension of the parameter and $n$ is the sample size \cite{Akaike1974AIC}. Moreover, its negative log marginal likelihood (a.k.a. free energy) has asymptotic expansion represented by $nS_n + (d/2) \log n + O_p(1)$, where $S_n$ is the empirical entropy \cite{Schwarz1978BIC}.

\begin{figure}[t]
\begin{minipage}{0.5\hsize}
\begin{flushleft}
\includegraphics[width=7.5cm]{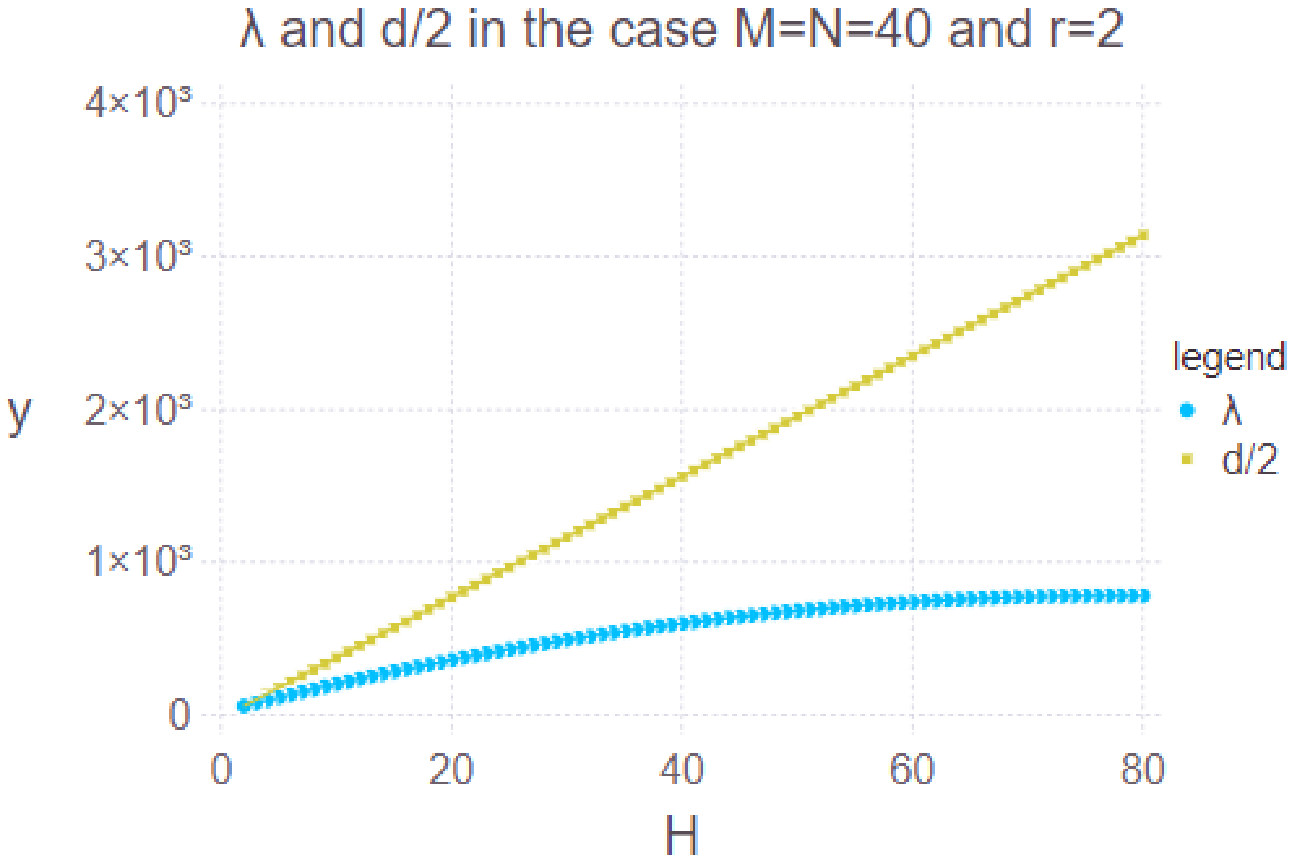}
\subcaption{}
\label{fig:RLCT}
\end{flushleft}
\end{minipage}
\begin{minipage}{0.5\hsize}
\begin{flushright}
\includegraphics[width=7.5cm]{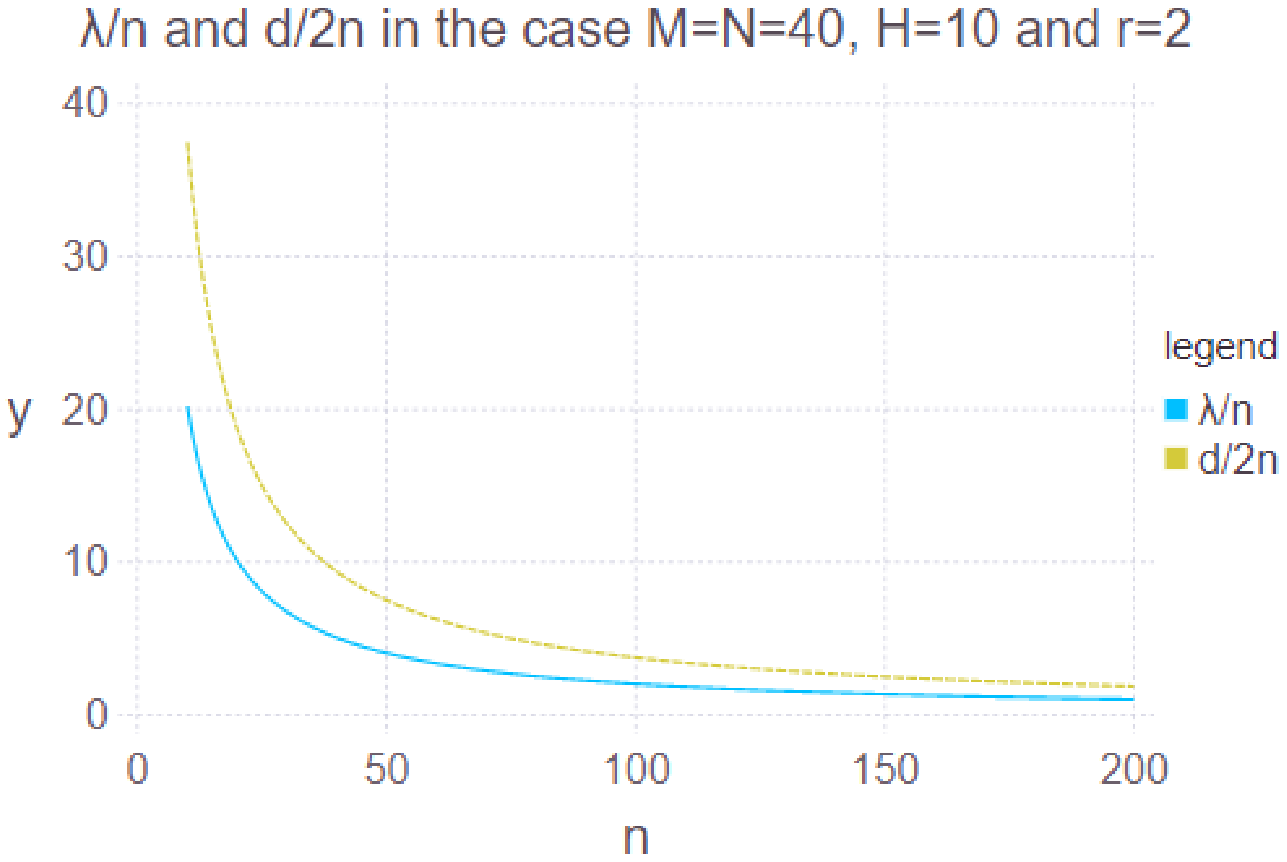}
\subcaption{}
\label{fig:LearningCurve}
\end{flushright}
\end{minipage}
\caption{(a) In this paper, we give the exact value of the learning coefficient of LDA $\lambda$.
The learning coefficient is smaller than half of the parameter dimension $d/2$, since LDA is a singular statistical model.
The dotted blue line drawn by the circles in this figure represents the learning coefficients of LDA when the number of topics $H$ is increased.
If LDA was a regular statistical model, its learning coefficient would be the dotted yellow line drawn by the squares.
The behavior of them are so different.
\\ (b) This figure shows the theoretical learning curve of LDA and that of a regular statistical model whose parameter dimension $d$ is same as LDA. The former is the solid blue line and the latter is the dashed yellow line.
The vertical axis means the expected generalization error $\mathbb{E}[G_n]$ and the horizontal one is the sample size $n$.
This is based on Eq. (\ref{thm-watanabeG}) and the exact value of $\lambda$ which is clarified by our result.}
\end{figure}
On the other hand, in the general case, by using resolution of singularity \cite{Hironaka}, Watanabe had proved that the asymptotic forms of its generalization error $G_n$ and marginal likelihood $Z_n$ are the followings \cite{Watanabe1,Watanabe2,SWatanabeBookE}:
\begin{align}
\label{thm-watanabeG}
\mathbb{E}[G_n] &= \frac{\lambda}{n} - \frac{m-1}{n \log n} + o\left(\frac{1}{n \log n} \right), \\
\label{thm-watanabeF}
-\log Z_n &= nS_n + \lambda \log n - (m-1) \log\log n + O_p(1),
\end{align}
where $\lambda$ is a positive rational number, $m$ is a positive integer, and $\mathbb{E}[\cdot]$ is an expectation operator on the overall datasets. The constant $\lambda$ is called a learning coefficient since it is dominant in the leading term of the above forms which represent learning curves (Fig. \ref{fig:LearningCurve}). The above forms hold not only in the case the model is regular but also in the case that is singular. For example, neural networks are singular statistical models and these generalization errors can be represented by Eq. (\ref{thm-watanabeG}) \cite{Watanabe2}. LDA is also singular; thus, this study is one of researches which clarify the generalization errors of singular statistical models. 
In the regular case, $\lambda=d/2$ and $m=1$ hold. However, in the singular case, they are depend on the model.
Let $K(w)$ be the Kullback-Leibler (KL) divergence between the true distribution to the statistical model, where $w$ is a parameter of the model. The function $K(w)$ is non-negative and analytic. The constants $\lambda$ and $m$ are characterized by a set of the zero points of the KL divergence: $K^{-1}(0)$. $K^{-1}(0)$ is an analytic set (a.k.a. algebraic variety). $\lambda$ is called a real log canonical threshold (RLCT) and $m$ is called a multiplicity in algebraic geometry. In LDA, if the number of topics changes, then $K^{-1}(0)$ does (Fig. \ref{fig:RLCT}).
A model selection method, called {\it sBIC} which uses RLCTs of statistical models, has been proposed by Drton and Plummer \cite{Drton}.
Drton and Imai have empirically verified that sBIC is precise to select the optimal and minimal model if the exact values or tight bounds of RLCTs are clarified \cite{Drton, Drton2017forest, Imai2019estimating}.
Other application of RLCTs is a design procedure for exchange probability in exchange Monte Carlo method by Nagata \cite{Nagata2008asymptotic}. To determine $\lambda$ and $m$, we should consider resolution of singularity for those concrete varieties. In general, we should find RLCTs to a family of functions to clarify a learning coefficient of a singular statistical model. There is no standard method to find RLCTs to a given collection of functions; thus, researchers study RLCTs with considering different procedures for each statistical model. In fact, RLCTs of several models has been analyzed in both statistics and machine learning. For instance, the RLCTs had been studied in Gaussian mixture model \cite{Yamazaki1}, Poisson mixture model \cite{SatoK2019PMM}, reduced rank regression \cite{Aoyagi1}, three-layered neural networks \cite{Watanabe2}, naive Bayesian networks \cite{Rusakov2005asymptotic}, Bayesian networks \cite{Yamazaki3}, Boltzmann machines \cite{Yamazaki4,Aoyagi2,Aoyagi3}, Markov models \cite{Zwiernik2011asymptotic}, hidden Markov models \cite{Yamazaki2}, Gaussian latent tree and forest models \cite{Drton2017forest}, and non-negative matrix factorization \cite{nhayashi2,nhayashi5,nhayashi8}. Note that clarifying the exact value of the RLCT in the all case is challenging problem. Whereas we would like to emphasize that this is not to deny the value and novelty of these researches, in deed, they cannot have clarified the exact value except for Aoyagi's result in 2005 \cite{Aoyagi1}.

In this paper, we derive the exact asymptotic form of the Bayesian generalization error by determination of the exact RLCT in LDA.
This article is divided to four parts. First, we introduced background of this research in the above. Second, we describe the framework of Bayesian inference and relationship between its theory and algebraic geometry. Third, we state the Main Theorem. Fourth, we report result of numerical experiment which is carried out in order to verify the behavior of Main Theorem if the sample size is finite. Lastly, we discuss about this theoretical result and we conclude this paper. We prove the Main Theorem in Appendix.

\section{Framework of Bayesian Inference and its Theory}
 
In this section, we briefly explain the framework of Bayesian inference and its mathematical theory.
Let $\mathcal{D} = (X_1, \ldots, X_n)$ be a sample (a.k.a. dataset: a collection of random variables) of $n$ independent and identically distributed from a data generating distribution (a.k.a. true distribution). The densities of the true distribution and a statistical model is denoted by $q(x)$ and $p(x|w)$, respectively. These domain $\mathcal{X}$ is a subset of a finite-dimensional real Euclidean or discrete space. Let $\varphi(w)$ be a probability density of a prior distribution. 
We define a posterior distribution as the following density function on $W$:
\begin{equation}
\label{def-posterior}
\varphi^*(w|\mathcal{D}) = \frac{1}{Z_n}\varphi(w) \prod_{i=1}^n p(X_i|w),
\end{equation}
where $Z_n$ is a normalizing constant to satisfy the condition $\int \varphi^*(w|\mathcal{D})dw=1$:
\begin{equation}
\label{def-marginallikelihood}
Z_n =  \int \mathrm{d}w \varphi(w) \prod_{i=1}^n p(X_i|w).
\end{equation}

This is called a marginal likelihood or a partition function. Its negative log value is called a free energy $F_n = -\log Z_n$.
Note that the marginal likelihood is a probability density function of a dataset. 

The free energy appears in a leading term of the difference between the true distribution and the model in the sense of dataset generating process. 
An entropy of the true distribution and an empirical one are denoted by
\begin{align}
\label{def-entropy}
S &= -\int \mathrm{d}x q(x) \log q(x), \\
S_n &= -\frac{1}{n} \sum_{i=1}^n \log q(X_i).
\end{align}
The KL divergence between the true distribution to the statistical model is denoted by
\begin{equation}
\label{def-aveerror}
K(w) = \int \mathrm{d}x q(x) \log \frac{q(x)}{p(x|w)}.
\end{equation}
By definition, $X_i \sim q(x)$ for $i=1,\ldots,n$ and $\mathcal{D} \sim \prod_{i=1}^n q(x_i)$ hold; thus let $\mathbb{E}_n[\cdot]$ be an expectation operator on overall dataset defined by
\begin{equation}
\label{def-expectation}
\mathbb{E}_n[\cdot] = \int \prod_{i=1}^n \mathrm{d}x_i q(x_i) [\cdot].
\end{equation}
The symbol $\mathbb{E}_n$ is referred to as $\mathbb{E}$ so long as there is no risk of confusion.
Obviously, $\mathbb{E}_n[S_n]$ is $S$.
Then, we have the following KL divergence
\begin{align}
\int \prod_{i=1}^n \mathrm{d}x_i q(x_i) \log \frac{\prod_{i=1}^n q(x_i)}{Z_n}
&= -\mathbb{E}_n\left[nS_n \right] - \mathbb{E}_n[\log Z_n] \\
&= -nS + \mathbb{E}_n[F_n].
\end{align}
The expected free energy is an only term depend on the model and the prior.
For this reason, the free energy is used as a criterion to select the model.

A predictive distribution is defined by the following density function on $\mathcal{X}$:
\begin{equation}
\label{def-predictive}
p^*(x | \mathcal{D} ) = \int \mathrm{d}w \varphi^*(w|\mathcal{D}) p(x|w).
\end{equation}
This is a probability distribution of a new data. It is also important for statistics and machine learning to evaluate the dissimilarity between the true and the model in the sense of a new data generating process.
A generalization error $G_n$ is defined by a KL divergence between the true distribution and the predictive one:
\begin{equation}
\label{def-gerror}
G_n = \int \mathrm{d}x q(x) \log \frac{q(x)}{p^*(x|\mathcal{D})}.
\end{equation}

Bayesian inference is defined by inferring that the true distribution may be the predictive one.
For an arbitrary finite $n$, by the definition of the marginal likelihood (\ref{def-marginallikelihood})
and the predictive distribution (\ref{def-predictive}), we have
\begin{align}
p^*(X_{n+1}|\mathcal{D}) &= \frac{1}{Z_n}\int \mathrm{d}w \varphi(w) \prod_{i=1}^n p(X_i|w) p(X_{n+1}|w) \\
&= \frac{1}{Z_n} \int \mathrm{d}w \varphi(w) \prod_{i=1}^{n+1} p(X_i|w) \\
&= \frac{Z_{n+1}}{Z_n}.
\end{align}
Thus, the following equation holds:
\begin{align}\label{Eq-GandFrelation}
-\log p^*(X_{n+1}|\mathcal{D}) &= -\log Z_{n+1} - (-\log Z_n) \\
&= F_{n+1} - F_n.
\end{align}
We can treat the expectation by $X_{n+1} \sim q(x_{n+1})$ as $\int \mathrm{d}x q(x)[\cdot]$ since the sample $(\mathcal{D},X_{n+1})$ is independent. Hence, we have
\begin{equation}
\int \mathrm{d}x_{n+1} q(x_{n+1}) [\cdot]
= \int \mathrm{d}x q(x) [\cdot].
\end{equation}
By definition, 
\begin{align}
&\quad -\int \mathrm{d}x_{n+1} q(x_{n+1}) \log p^*(x_{n+1}|\mathcal{D}) \\
&= \int \mathrm{d}x_{n+1} q(x_{n+1}) \log \frac{q(x_{n+1})}{p^*(x_{n+1}|\mathcal{D})} - \int \mathrm{d}x_{n+1} q(x_{n+1}) \log  q(x_{n+1}) \\
&=G_n + S
\end{align}
holds; thus, with integrating the both sides in Eq. (\ref{Eq-GandFrelation}) by $\mathrm{d}x_{n+1} q(x_{n+1})$, we obtain
\begin{equation}
G_n + S = \int \mathrm{d}x_{n+1} q(x_{n+1}) F_{n+1} - F_n.
\end{equation}
Considering expectation on overall dataset, we get
\begin{align}
\mathbb{E}_{n}[G_n] + S &= \mathbb{E}_{n+1}[F_{n+1}] - \mathbb{E}_{n}[F_n].
\end{align} 
Hence, $G_n$ and $F_n$ are important random variables in Bayesian inference.
In mathematical theory of Bayesian statistics (a.k.a. singular learning theory),
we consider how they asymptotically behave in the general case \cite{SWatanabeBookMath}.
To establish this theory, resolution of singularity in algebraic geometry has been needed.

Now, we briefly explain the relationship between singular learning theory and algebraic geometry.
As technical assumptions, we suppose the parameter set $W \subset \mathbb{R}^d$ is sufficiently wide compact and the prior is positive and bounded on $K^{-1}(0)$: $0<\varphi(w)<\infty$ for any $w \in K^{-1}(0)$. Moreover, $\varphi(w)$ is a $C^{\infty}$-function on with the compact support $W$.
Considering $K(w)$ in Eq. (\ref{def-aveerror}) and its zero points $K^{-1}(0)$,
we use the following analytic form by \cite{Atiyah1970resolution} of the singularities resolution theorem \cite{Hironaka}. 
Atiyah has derived this form of the singularities resolution theorem in order to analyze the relationship between a division of distributions (hyperfunctions) and local type zeta functions \cite{Atiyah1970resolution}.
Watanabe has proved that it is useful for constructing singular learning theory \cite{Watanabe1, Watanabe2, SWatanabeBookE}.
\begin{thm}[Singularities Resolution Theorem]\label{thm-singular}
Let $H$ be a non-negative analytic function on $W \subset \mathbb{R}^d$.
Assume that $H^{-1}(0)$ is not an empty set.
Then, there are an open set $W'$, a $d$-dimensional smooth manifold $\mathcal{M}$, and an analytic map $g:\mathcal{M} \rightarrow W'$ such that $g: \mathcal{M}\setminus g^{-1}(H^{-1}(0)) \rightarrow W'\setminus H^{-1}(0)$ is isomorphic and
\begin{gather*}
H(g(u))=u_1^{2k_1} \ldots u_d^{2k_d}, \\
|\det g'(u)| =b(u)|u_1^{h_1} \ldots u_d^{h_d}|
\end{gather*}
hold for each local chart $U \ni u$ of $\mathcal{M}$,
where $k_j$ and $h_j$ are non-negative integer for $j=1,\ldots,d$, $\det g'(u)$ is the Jacobian of $g$ and $b:\mathcal{M} \rightarrow \mathbb{R}$ is strictly positive analytic: $b(u)>0$. 
\end{thm}
Let $\mathrm{Re}(z)$ be the real part of a complex number $z$: $\mathrm{Re}(\alpha+\beta\sqrt{-1})=\alpha$, where $\alpha$ and $\beta$ are real numbers and $\sqrt{-1}$ is the imaginary unit. By using Theorem \ref{thm-singular},
the following theorem is proved \cite{Atiyah1970resolution, Bernstein1972, Sato1974zeta}.
\begin{thm}\label{thm-zeta}
Let $F: \mathbb{R}^d \rightarrow \mathbb{R}$ be an analytic function of a variable $w \in \mathbb{R}^d$.
$a: W \rightarrow \mathbb{R}$ is denoted by a $C^{\infty}$-function with compact support $W$. Then
$$\zeta(z) = \int_W |F(w)|^z a(w) \mathrm{d}w$$
is a holomorphic function in $\mathrm{Re}(z)>0$. Moreover, $\zeta(z)$ can be analytically continued to a unique meromorphic
function on the entire complex plane $\mathbb{C}$. Its all poles are negative rational numbers.
\end{thm}
\begin{example}
As a simple example of Theorem \ref{thm-zeta}, put $F(\theta)=\theta^2$ and $a(\theta)=1$, where $\theta \in [0,1]$.
Consider the univariate function of a complex number $z$
\begin{equation*}
\zeta(z) = \int_0^1 \theta^{2z} \mathrm{d} \theta.
\end{equation*}
If $\mathrm{Re}(z)$>0, then this integral is immediately calculated as
\begin{equation*}
\int_0^1 \theta^{2z} \mathrm{d} \theta = \frac{1}{2z+1}.
\end{equation*}
Hence,  $\zeta(z)$ is a holomorphic function in $\mathrm{Re}(z)>0$.
However, $z \mapsto 1/(2z+1)$ is a meromorphic function on the entire complex plane and its pole is $z=-1/2$.
Therefore, we can analytically continue $\zeta(z)$ as the meromorphic function whose pole is $z=-1/2$.
\end{example} 
Applying Theorem \ref{thm-singular} to the KL divergence in Eq. (\ref{def-aveerror}), we have
\begin{gather}
K(g(u))=u_1^{2k_1} \ldots u_d^{2k_d}, \\
|\det g'(u)| =b(u)|u_1^{h_1} \ldots u_d^{h_d}|.
\end{gather}
Suppose the prior density $\varphi(w)$ has the compact support $W$ and the open set $W'$ satisfies $W \subset W'$.
By using Theorem \ref{thm-zeta}, we can define a zeta function of learning theory.

\begin{defi}[Zeta Function of Learning Theory]
\label{def-learnzeta}
Let $K(w) \geqq 0$ be the KL divergence mentioned in Eq. (\ref{def-aveerror}) and $\varphi(w)\geqq 0$ be a prior density function which satisfies the above assumption. A zeta function of learning theory is defined by the following univariate complex function
$$\zeta(z) = \int_W K(w)^z \varphi(w) \mathrm{d}w.$$
\end{defi}

\begin{defi}[Real Log Canonical Threshold]
Let $\zeta(z)$ be a zeta function of learning theory represented in Definition \ref{def-learnzeta}.
Consider an analytic continuation of $\zeta(z)$ from Theorem \ref{thm-zeta}.
A real log canonical threshold (RLCT) is defined by the negative maximum pole of $\zeta(z)$ and its multiplicity is defined by the order of the maximum pole. 
\end{defi}

Here, we describe how to determine the RLCT $\lambda>0$ of the model corresponding to $K(w)$.
We apply Theorem \ref{thm-singular} to the zeta function of learning theory.
For each local coordinate $U$, we have
\begin{align}
\zeta(z) &= \int_U K(g(u))^z \varphi(g(u)) |\det g'(u)| \mathrm{d}u \\
&= \int_U u_1^{2k_1z+h_1} \ldots u_d^{2k_dz+h_d}\varphi(g(u))b(u) \mathrm{d}u.
\end{align}
The functions $\varphi(g(u))$ and $b(u)$ are strictly positive in $U$;
thus, we should consider the maximum pole of
\begin{equation}
\int_U u_1^{2k_1z+h_1} \ldots u_d^{2k_dz+h_d} \mathrm{d}u = \frac{C_1(z)}{2k_1 z + h_1}\ldots\frac{C_d(z)}{2k_d z + h_d},
\end{equation}
where $(C_j(z))_{j=1}^d$ are non-zero functions of $z \in \mathbb{C}$. Hence, we give the RLCT in the local chart $U$ as follows
\begin{equation}
\lambda_U = \min_{j=1}^d \left\{ \frac{h_j + 1}{2k_j} \right\}.
\end{equation}
By considering the duplication of indices, we can also find the multiplicity $m$.
Therefore, we can determine the RLCT as $\lambda = \min_U \lambda_U$.
The RLCT is equal to the learning coefficient because of Eq. (\ref{thm-watanabeG}) and (\ref{thm-watanabeF}).
That is why we need resolution of singularity to clarify the behavior of the Bayesian generalization error $G_n$ and the free energy $F_n$ via determination of the learning coefficient.

\section{Main Theorem}
\label{sec-main}
In this section, we state the Main Theorem: the exact value of the RLCT of LDA. 
\begin{defi}[Stochastic Matrix]
A stochastic matrix is defined by a matrix whose columns are in simplices,
i.e. the sum of the entries in each column is equal to one and each entries are non-negative.

An $M \times N$ matrix $C=(c_{ij})_{i=1,j=1}^{M,N}$ is stochastic if and only if $c_{ij} \geqq 0$ and $\sum_{j=1}^N c_{ij} = 1$ hold.
\end{defi}

In the following, the parameter $w=(A,B)$ is a pair of stochastic matrices and the data $x$ is a one-hot vector.
By definition, a set of stochastic matrices is compact. Let $\mathrm{Onehot}(N):=\{w = (w_j) \in \{0,1\}^N \mid \sum_{j=1}^N w_j =1\}$ be a set of $N$-dimensional one-hot vectors and
$\mathrm{\Delta}_N:=\{c=(c_j)_{j=1}^N \mid \sum_{j=1}^N c_j=1\}$ be an $N$-dimensional simplex.
In LDA terminology, the number of documents and the vocabulary size is denoted by $N$ and $M$, respectively.
Let $H_0$ be the optimal (or true) number of topics and $H$ be the chosen one. In this situation, the sample size $n$ is the number of words in all of the given documents.
Note that $x_i$ is the $i$-th entry of the one-hot vector $x$ in this section (it is not the realization of the $i$-th data $X_i$ of the dataset like the previous section).
See also Table \ref{params} (this table is quoted and modified from our previous study \cite{nhayashi7}).

\begin{table}[htb]
  \centering
  \caption{Description of Variables in LDA Terminology}
  \begin{tabular}{|c|c|c|} \hline
    Variable & Description & Index \\
    \hline\hline
    $b_j=(b_{kj}) \in \mathrm{\Delta}_H$ & topic proportion of topic $k$ in document $j$ & for $k=1,\ldots,H$ \\
    $a_k =(a_{ik}) \in \mathrm{\Delta}_M$ & appearance probability of word $i$ in topic $k$ & for $i=1,\ldots,M$ \\
    \hline
    $x=(x_i) \in \mathrm{Onehot}(M)$ & word $i$ is defined by $x_i=1$ & for $i=1,\ldots,M$  \\
    $y=(y_k) \in \mathrm{Onehot}(H)$ & topic $k$ is defined by $y_k=1$ & for $k=1,\ldots,H$  \\
    $z=(z_j) \in \mathrm{Onehot}(N)$ & document $j$ is defined by $z_j=1$ & for $j=1,\ldots,N$  \\
    \hline
    $*_0$ and $*^0$ & optimal or true variable corresponding to $*$ & - \\
  \hline
  \end{tabular}
\label{params}
\end{table}

Let $A=(a_{ik})_{i=1,k=1}^{M,H}$ and $B=(b_{kj})_{k=1,j=1}^{H,N}$ be $M \times H$ and $H \times N$ stochastic matrix, respectively.
Assume that the pair of stochastic matrices $(A_0, B_0)$ is one of optimal parameters, where $A_0=(a^0_{ik})_{i=1,k=1}^{M,H_0}$ and $B_0=(b^0_{kj})_{k=1,j=1}^{H_0,N}$. Suppose $A_0$ and $B_0$ are full rank.

Here, we define the RLCT of LDA in the below. This definition is also quoted and modified from the statement in \cite{nhayashi7}.

\begin{defi}[LDA]
\label{def-LDAmodel}
Assume that $M\geqq 2$, $N \geqq 2$, and $H \geqq H_0 \geqq 1$.
Let $q(x|z)$ and $p(x|z,A,B)$ be conditional probability mass functions of $x \in \mathrm{Onehot}(M)$ given $z \in \mathrm{Onehot}(N)$ as the following:
\begin{align}
\label{topic-true}
q(x|z) & = \prod_{j=1}^N \left(\sum_{k=1}^{H_0} b^0_{kj} \prod_{i=1}^M (a^0_{ik})^{x_{i}}\right)^{z_j}, \\
\label{topic-model}
p(x|z,A,B) & = \prod_{j=1}^N \left(\sum_{k=1}^{H} b_{kj} \prod_{i=1}^M (a_{ik})^{x_{i}}\right)^{z_j}. 
\end{align}
The prior density function is denoted by $\varphi(A,B)$.
The conditional mass $q(x|z)$ and $p(x|z,A,B)$ represent the true distribution of LDA and the statistical model of that, respectively.
\end{defi}

These distributions are the marginalized ones of the followings which contain the true topics $y^0 \in \mathrm{Onehot}(H_0)$ and the one of the model $y \in \mathrm{Onehot}(H)$:
\begin{align*}
q(x,y^0|z) & = \prod_{j=1}^N \left[\prod_{k=1}^{H_0} \left(b^0_{kj} \prod_{i=1}^M (a^0_{ik})^{x_{i}}\right)^{y^0_k}\right]^{z_j}, \\
p(x,y|z,A,B) & = \prod_{j=1}^N \left[\prod_{k=1}^{H} \left(b_{kj} \prod_{i=1}^M (a_{ik})^{x_{i}}\right)^{y_k}\right]^{z_j}. 
\end{align*}
In fact, $q(x|z) = \sum_{y^0 \in \mathrm{Onehot}(H_0)} q(x, y^0 | z)$ and $p(x|z, A, B) = \sum_{y \in \mathrm{Onehot}(H_0)} p(x,y|z,A,B)$ hold because of $z \in \mathrm{Onehot}(N)$.
In practical cases, the topics are not observed; thus, we use Eqs. (\ref{topic-true}) and (\ref{topic-model}) as the definition of LDA.

\begin{defi}[RLCT of LDA]
\label{def-LDARLCT}
Let $K(A,B)$ be the KL divergence between $q(x|z)$ and $p(x|z,A,B)$:
$$K(A,B)=\sum_{z \in \mathrm{Onehot}(M)} \sum_{x \in \mathrm{Onehot}(N)} q(x|z)q'(z) \log \frac{q(x|z)}{p(x|z,A,B)},$$
where $q'(z)$ is the true distribution of the document. In LDA, $q'(z)$ is not observed and assumed that it is positive and bounded.
Assume that $\varphi(A,B) >0$ is positive and bounded on $K^{-1}(0) \ni (A_0,B_0)$.
Then, the zeta function of learning theory in LDA is
the holomorphic function of univariate complex variable $z \ (\mathrm{Re} (z) >0)$
\[
\zeta(z)=\iint
 K(A,B)^z \mathrm{d}A\mathrm{d}B.
\]
Because of Theorem \ref{thm-zeta}, $\zeta(z)$ can be analytically continued to a unique meromorphic function on the entire complex plane $(z \in \mathbb{C})$ and all of its poles are negative rational numbers.
The RLCT of LDA is defined by $\lambda$ if the largest pole of $\zeta(z)$ is $(-\lambda)$.
Its multiplicity $m$ is defined as the order of the maximum pole.
\end{defi}

In order to prepare to state Main Theorem,
we define an intrinsic value $r$ of the true distribution defined by stochastic matrices $A_0$ and $B_0$.
Let $\tilde{A}_0^{\setminus (M,H_0)}$ and $\tilde{B}_0^{\setminus (H_0, 1)}$ be matrices defined as
\begin{equation}
\tilde{A}_0^{\setminus (M,H_0)} = (a_{ik}^0)_{i=1, k=1}^{M-1, H_0-1}
\end{equation}
and
\begin{equation}
\tilde{B}_0^{\setminus (H_0, 1)} = (b_{kj}^0)_{k=1, j=2}^{H_0-1, N},
\end{equation}
respectively. Also, $\tilde{A}_0^{(\setminus M), H_0}$ and $\tilde{B}_0^{(\setminus H_0), 1}$ are denoted by matrices
whose column vectors are same such that
\begin{equation}
\tilde{A}_0^{(\setminus M), H_0} = (a^0_{(1:M-1)H_0}, \ldots a^0_{(1:M-1)H_0}) = (a^0_{iH_0})_{i=1,k=1}^{M-1, H_0-1}
\end{equation}
and
\begin{equation}
\tilde{B}_0^{(\setminus H_0), 1} = (b^0_{(1:H_0-1)N}, \ldots b^0_{(1:H_0-1)1}) = (b^0_{k1})_{k=1,l=1}^{H_0-1, N-1},
\end{equation}
where $a^0_{(1:M-1)H_0}=(a^0_{iH_0})_{i=1}^{M-1}$ and $b^0_{(1:H_0-1)1}=(b^0_{k1})_{k=1}^{H_0-1}$ are an ($M-1$)-dimensional vector and an ($H_0-1$)-dimensional vector, respectively.
Then, let
\begin{align}
U_0 &=\tilde{A}_0^{\setminus (M,H_0)} - \tilde{A}_0^{(\setminus M), H_0}, \label{defU0} \\
V_0 &= \tilde{B}_0^{\setminus (H_0, 1)} - \tilde{B}_0^{(\setminus H_0), 1}, \label{defV0}
\end{align}
and $r=\mathrm{rank} (U_0V_0)$. Obviously, $r$ depends on $H_0$.

The main result of this paper is the following theorem.

\begin{thm}[Main Theorem]
\label{thm-main}
Suppose $M\geqq 2$, $N \geqq 2$, and $H \geqq H_0 \geqq 1$.
Let $r$ be the rank of $U_0V_0$ which is a product of two matrices $(U_0, V_0)$ defined in Eqs. (\ref{defU0}) and (\ref{defV0}).
The RLCT of LDA $\lambda$ and its multiplicity $m$ are as follows.

\begin{enumerate}
\item If $N+r+1 \leqq M+H$ and $M+r+1\leqq N+H$ and $H+r+1\leqq M+N$,
    \begin{enumerate}
        \item and if $M+N+H+r$ is odd, then
        $$\lambda = \frac{1}{8}\{ 2(H+r+1)(M+N)-(M-N)^2-(H+r+1)^2 \} - \frac{1}{2}N, \ m = 1.$$
        \item and if $M+N+H+r$ is even, then
        $$\lambda = \frac{1}{8}\{ 2(H+r+1)(M+N)-(M-N)^2-(H+r+1)^2 +1 \} - \frac{1}{2}N, \ m = 2.$$
    \end{enumerate}
\item Else if $M+H<N+r+1$, then
$$\lambda = \frac{1}{2}\{ MH+N(r+1)-H(r+1)-N \}, \ m = 1.$$
\item Else if $N+H<M+r+1$, then
$$\lambda = \frac{1}{2}\{ NH+M(r+1)-H(r+1)-N \}, \ m = 1.$$
\item Else (i.e. $M+N<H+r+1$), then
$$\lambda = \frac{1}{2}(MN-N), \ m = 1.$$
\end{enumerate}
\end{thm}

To prove Main Theorem, we use the RLCT of matrix factorization (MF).

\begin{defi}[RLCT of MF]
Let $U$, $V$, $U_0$ and $V_0$ be $M \times H$, $H \times N$, $M \times H_0$ and $H_0 \times N$ real matrix, respectively.
Set $M \geqq 1$, $N \geqq 1$, $H \geqq H_0 \geqq 0$. Assume that they in a compact subset $W$ of $(M+N)H$-dimensional Euclidean space and put $r = \mathrm{rank}(U_0V_0)$.
The RLCT of MF $\lambda_{\mathrm{MF}} = \lambda_{\mathrm{MF}}(M,N,H,r)$ is defined by the negative maximum pole of the following zeta function
$$\zeta_{\mathrm{MF}}(z) = \iint_W \lVert UV -U_0V_0 \rVert^{2z} \mathrm{d}U\mathrm{d}V.$$
Its multiplicity $m_{\mathrm{MF}}$ is defined as the order of the maximum pole.
\end{defi}

The exact value of $\lambda_{\mathrm{MF}}$ had been clarified as that of reduced rank regression (a.k.a. three-layered linear neural network) by Aoyagi in \cite{Aoyagi1}. By making the RLCT of LDA come down to that of MF, we prove Main Theorem.
The rigorous proof of Main Theorem is described in Appendix.

\begin{proof}[Sketch of Proof]
Let $\zeta_{\mathrm{SMF}}(z)$ be a zeta function of learning theory in stochastic MF:
\begin{equation}
\zeta_{\mathrm{SMF}}(z) = \iint \lVert AB - A_0B_0 \rVert^{2z} \mathrm{d}A\mathrm{d}B.
\end{equation}
Let $\lambda_{\mathrm{SMF}}$ and $m_{\mathrm{SMF}}$ be the negative maximum pole and its order of
$\zeta_{\mathrm{SMF}}(z)$, respectively.
According to \cite{Matsuda1-e} and \cite{nhayashi7}, there exists two positive constants $C_1>0$ and $C_2>0$ such that 
\begin{equation}
C_1 K(A,B) \leqq \lVert AB - A_0B_0 \rVert^2 \leqq C_2 K(A,B),
\end{equation}
i.e. 
\begin{gather}
\lambda = \lambda_{\mathrm{SMF}}, \
m = m_{\mathrm{SMF}}
\end{gather}
hold; thus, we only have to consider $\lambda_{\mathrm{SMF}}$ and $m_{\mathrm{SMF}}$.

Developing $\lVert AB - A_0B_0 \rVert^2$,
performing several changes of variables
and considering the integral range of transformed variables in the zeta function,
we have
\begin{gather}
\lambda_{\mathrm{SMF}} = \frac{M-1}{2} + \lambda_{\mathrm{MF}}(M-1,N-1,H-1,r), \
m_{\mathrm{SMF}} = m_{\mathrm{MF}}.
\end{gather}

We calculate $\lambda_{\mathrm{MF}}$ and $m_{\mathrm{MF}}$ by following \cite{Aoyagi1},
then we obtain Main Theorem.
\end{proof}

\section{Experiment}

Now, we run numerical experiments to check the behavior of Main Theorem when the sample size is finite.
Theorem \ref{thm-main} gives the exact asymptotic form of Bayesian generalization error in LDA by using Eq. (\ref{thm-watanabeG}).
We calculate the Bayesian generalization error in LDA by using Gibbs sampling and compare the numerically-calculated RLCT with the theoretical one. Our experimental approach and its description in this section is based on \cite{nhayashi8} since it also treats the numerical experiment to compute the RLCT by using Gibbs sampling to verify the numerical behavior of its theoretical result. 

Let $\mathbb{E}_w$ be an expectation operator of the posterior: $\mathbb{E}_w[\cdot]=\int \mathrm{d}w \varphi^*(w|\mathcal{D})[\cdot]$. Let $\hat{\lambda}$ be the numerically calculated RLCT.
The widely applicable information criterion (WAIC) \cite{WatanabeAIC} is defined by the following random variable $W_n$:
\begin{equation}
W_n = T_n + V_n / n,
\end{equation}
where $T_n$ is the empirical loss and $V_n$ is the functional variance:
\begin{align}
T_n &= -\frac{1}{n} \sum_{i=1}^n p^*(X_i) = -\frac{1}{n} \sum_{i=1}^n \mathbb{E}_w [p(X_i | w)], \\
V_n &= \sum_{i=1}^n \left[ \mathbb{E}_w [(\log p(X_i|w))^2] - \left\{ \mathbb{E}_w [ \log p(X_i|w)] \right\}^2 \right]
= \sum_{i=1}^n \mathbb{V}_w[\log p(X_i | w)].
\end{align}
Even if the posterior distribution cannot be approximated by any normal distribution (i.e., the model is singular),
the expected WAIC $\mathbb{E}[W_n]$ is asymptotically equal to the expected generalization loss $\mathbb{E}[G_n + S]$ \cite{WatanabeAIC};
\begin{equation}
\mathbb{E}[G_n + S] = \mathbb{E}[W_n] + o(1/n^2).
\end{equation}
Moreover, the generalization error and the WAIC error $W_n - S_n$ have the same variance \cite{WatanabeAIC, SWatanabeBookMath}:
\begin{align}\label{var_RLCT}
G_n + W_n - S_n = 2\lambda /n + o_p(1/n).
\end{align}
We need to repeat the simulations to compute $\hat{\lambda}$ to decrease the random effect caused by $G_n$, $W_n$ and $S_n$.
Thus, Eq. (\ref{var_RLCT}) is useful for computing $\hat{\lambda}$ because the leading term $2\lambda /n$ is deterministic, nevertheless the left hand side is probabilistic. This means that the needed number of simulations $D$ can be less than that in the case using Eq. (\ref{thm-watanabeG}).

The method was as follows.
First, the training data $\mathcal{D}$ was generated from the true distribution $q(X|Z)$.
Second, the posterior distribution was calculated by using Gibbs sampling \cite{Griffiths2004LDAGS}.
Third, $G_n$ and $W_n - S_n$ were computed by using the training data $\mathcal{D}$ and the artificial test data $\mathcal{D^*}=(X_t^*)$ generated from $q(X|Z)$.
These three steps were repeated and each value of  $n(G_n + W_n - S_n)/2$ was saved.
After all repetitions have been completed,
$n(G_n + W_n - S_n)/2$ was averaged over the simulations. The average value was $\hat{\lambda}$.

The pseudo-code is listed in Algorithm \ref{exp_alg}, where $K$ is the sample size of the parameter subject to the posterior and $n_T$ is the sample size of the synthesized test data.
We used the programing language named Julia \cite{Bezanson2017julia} for this experiment.

\begin{algorithm}
\caption{How to Compute $\hat{\lambda}$}
\label{exp_alg}
\begin{algorithmic}
\REQUIRE $D$: the number of simulations, \\ 
$A_0$: the true parameter matrix whose size is $(M, H_0)$,\\
$B_0$: the true parameter matrix whose size is $(H_0, N)$,\\
$\mathrm{GS}$: the Gibbs sampling function whose return value consists of the samples from the posterior.
\ENSURE The numerical computed RLCT $\hat{\lambda}$.
\STATE Allocate an array $\Lambda[D]$.
\FOR{$d=1$ to $D$}
\STATE Generate $\mathcal{D}$ from the true distribution.
\STATE Allocate arrays $\mathcal{A}[M, H, K]$ and $\mathcal{B}[H, N, K]$.
\STATE Get $\mathcal{A}, \mathcal{B} \leftarrow \mathrm{GS}(\mathcal{D})$.
\STATE Generate $\mathcal{D^*}$ from the true distribution.
\STATE Calculate $G_n \approx \frac{1}{n_T} \sum_{t=1}^{n_T} \log \frac{q(X^*_t)}{\mathbb{E}_w [p(X^*_t | w)]}$,
$S_n = -\frac{1}{n} \sum_{i=1}^n \log q(X_i)$,
\STATE and $W_n \approx -\frac{1}{n} \sum_{i=1}^n \mathbb{E}_w [p(X_i | w)] + \frac{1}{n} \sum_{i=1}^n \mathbb{V}_w[\log p(X_i | w)]$,
\STATE $\quad$ where $\mathbb{E}_w[ f(w) ] \approx \frac{1}{K} \sum_{k=1}^K f(w_k)$ and $w_k = (\mathcal{A}[:,:,k], \mathcal{B}[:,:,k])$.
\STATE Save $\Lambda[d] \leftarrow n(G_n + W_n - S_n)/2$.
\ENDFOR
\STATE Calculate $\hat{\lambda} = \frac{1}{D} \sum_{d=1}^D \Lambda[d]$.
\end{algorithmic}
\end{algorithm}


We set 
$M=10$, $N=5$, $H_0=2$, $r=1$, $n=1000$, and $n_T = 200n = 200000$.
To examine the numerical behavior when the number of topics $H \geqq H_0$ in the model increases, we set $H=2,3,4,5$ and carried out experiments in each case.
To decrease the probabilistic effect of Eq.~(\ref{var_RLCT}), we conducted the simulations one hundred times, i.e. $D=100$.

In the Gibbs sampling, we had to conduct a burn-in to decrease the effect of the initial values and thin the samples in order to break the correlations.
The length of the burn-in was $10000$, while the length of the thinning was $20$; thus, the sample sizes of the parameter was $20000 + 20K = 50000$ ($K=1000$) and we used the $(10000 + 20k)$-th sample as the entry of $\mathcal{A}[:, :, k]$ and $\mathcal{B}[:, :, k]$ for $k=1$ to $K$.



\begin{figure}[h]
\centering
\includegraphics[width=11cm]{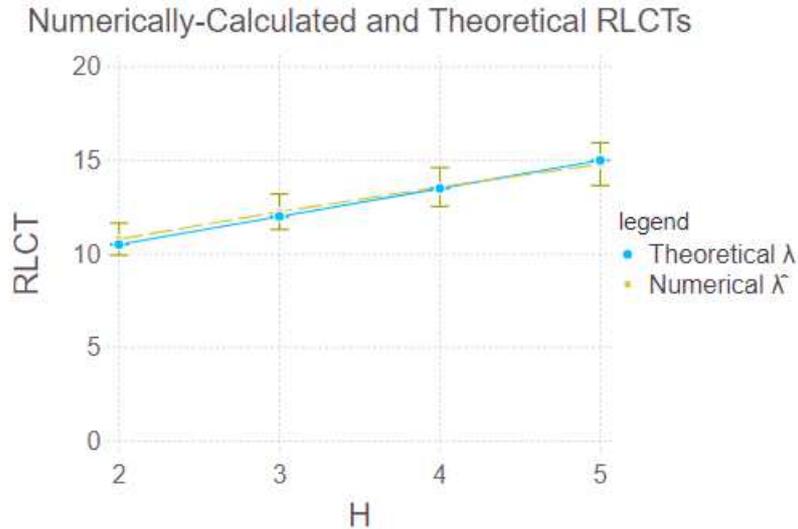}
\caption{This figure is drawn based on Table \ref{exp-result} and Theorem~\ref{thm-main}. It compares numerically-calculated RLCTs $\hat{\lambda}$ (Numerical $\lambda^{\hat{}}$, as the dashed yellow line with the error bars) and theoretical ones $\lambda$ (Theoretical $\lambda$, as the solid blue line) for $H=2,3.4.5$. The horizontal line means the number of topics $H$ and the vertical one is the numerically-calculated or theoretical value of the RLCT. Each error bar of experimental results is the 1-standard deviation range. The line of Numerical $\lambda^{\hat{}}$ and that of Theoretical $\lambda$ are very close and the standard deviations are sufficiently smaller than the scale of the RLCTs.}
\label{fig:exp-visual}
\end{figure}

The experimental results are shown in Table~\ref{exp-result} and visualized in Fig.~\ref{fig:exp-visual}.
The symbol $\lambda$ denotes the exact value of the RLCT $\lambda$ in Theorem~\ref{thm-main}.
There are three columns for each $H$, and each row contains the model settings (Settings), symbols of calculated values (RLCTs), and the theoretical or numerical values (Values). The experimental values have four significant digits.
The numerically-calculated RLCT $\hat{\lambda}$ is an average of $n(G_n + W_n - S_n)/2$ obtained from each simulations;
hence, we also show the standard deviation of it as the right next to the plus-minus sign $\pm$ in each setting.


\begin{table}[h]\centering
\caption{Numerically-Calculated and Theoretical Values of RLCTs}
  \begin{tabular}{|c|c|c|c|} \hline
   \multicolumn{1}{|l|}{Settings}
      & \multicolumn{1}{l|}{RLCTs}
      & \multicolumn{1}{l|}{Values} \\ \hline
    $H=2$ &  Theoretical: $\lambda$ & 21/2 \\ 
    $(M=10,N=5$ & Numerical: $\hat{\lambda}$ & $10.79 \pm 0.8591$ \\ 
    $H_0=2,r=1)$ & Difference: $|\lambda-\hat{\lambda}|$ & $0.2901 \pm 0.8591$ \\ \hline
    $H=3$ &  Theoretical: $\lambda$ & 12 \\ 
    $(M=10,N=5$ & Numerical: $\hat{\lambda}$ & $12.25 \pm 0.9510$ \\ 
    $H_0=2,r=1)$ & Difference: $|\lambda-\hat{\lambda}|$ & $0.2534 \pm 0.9510$ \\ \hline
    $H=4$ &  Theoretical: $\lambda$ & 27/2 \\ 
    $(M=10,N=5$ & Numerical: $\hat{\lambda}$ & $13.57 \pm 1.036$ \\ 
    $H_0=2,r=1)$ & Difference: $|\lambda-\hat{\lambda}|$ & $0.07114 \pm 1.036$ \\ \hline
    $H=5$ &  Theoretical: $\lambda$ & 15 \\ 
    $(M=10,N=5$ & Numerical: $\hat{\lambda}$ & $14.80 \pm 1.143$ \\ 
    $H_0=2,r=1)$ & Difference: $|\lambda-\hat{\lambda}|$ & $0.2049 \pm 1.143$ \\ \hline
  \end{tabular}\label{exp-result}
\end{table}

As shown in Table~\ref{exp-result} and Fig.~\ref{fig:exp-visual}, all numerically calculated values are nearly equal to the theoretical RLCTs, i.e. these differences are sufficiently smaller than the standard deviation overall simulations.
Note that the parameter dimensions $(M-1)H+(H-1)N$ are $23$, $37$, $51$ and $65$ for $H=2$, $3$, $4$ and $5$;
thus, we consider that the sample size $n=1000$ is not an asymptotic scale. Moreover, we also consider that it is natural to fix the sample size while the number of topics increases because we compare some models for a dataset (the sample size is fixed) in practical situations.
Although the sample size is finite (not an asymptotic scale) and fixed, the theoretical values are included in the 1-standard deviation ranges for each case. Besides, because of Fig.~\ref{fig:exp-visual}, the standard deviations are sufficiently small for the scale of the RLCTs.
Therefore, Theorem~\ref{thm-main} is consistent with the experimental result.

\section{Discussion and Conclusion}
In this paper, we described how the exact RLCT (i.e. learning coefficient) of LDA is determined in the general case.
Main Theorem 
Using this result, we also clarified the exact asymptotic forms of the Bayesian generalization error and the marginal likelihood in LDA.

The RLCT of LDA can be represented by using that of MF.
Namely, Main Theorem can be interpreted as that the learning coefficient of LDA is that of the unconstrained MF minus the penalty due to the simplex constraint. In fact, it can be proved that
\begin{equation}
\lambda = \lambda_{\mathrm{MF}}(M,N,H,r+1)-\frac{N}{2}
\end{equation}
holds (see also the rigorous proof of Main Theorem in Appendix).
The dimension of the stochastic matrix $AB$ with the degrees of freedom is $(M-1)N=MN-N$.
The subtracted $N$ is the dimension of the parameter that can be uniquely determined from the parameters of the other $(M-1)N$ dimensions in the matrix $AB$. This part can be regarded as an N-dimensional regular statistical model, whose RLCT is $N/2$. This is the reason of the above statement.
Note that Main Theorem and its proof are not trivial. 
A hermeneutic explanation cannot be a mathematical proof.
In addition, the actual parameter dimension is $(M-1)H+(H-1)N=(M+N-1)H-N$ because we have to consider the matrices $A$ and $B$ rather than $AB$. We cannot reach the result of this paper simply by maintaining consistency of the degrees of freedom.
Algebraic geometrical methods are used to solve this problem in learning theory: what the learning coefficient of LDA is.

Since LDA is a knowledge discovery method, marginal-likelihood-based model selection often tends to be preferred.
However, BIC \cite{Schwarz1978BIC} cannot be used for LDA because it is a singular statistical model.
Although Gibbs sampling is usually used for full Bayesian inference of LDA, it is difficult to achieve a tempered posterior distribution; thus, we need other Markov chain Monte Carlo method (MCMC) to calculate WBIC \cite{WatanabeBIC} and WsBIC \cite{Imai2019estimating}. 
The result of this study allows us to perform a rigorous model selection of LDA with sBIC \cite{Drton}, which is MCMC-free. Even when the marginal likelihoods are computed directly by the exchange Monte Carlo method, our result is useful for the design of the exchange probability \cite{Nagata2008asymptotic}.
Furthermore, it may be possible to evaluate how precise MCMC approximates the posterior with the exact values of that \cite{SWatanabeBookMath, Imai2019estimating}.

One may use BIC for model selection of LDA; however, using it causes that too small models are chosen.
This is because there exists a large difference in values and behaviors between $d/2$ and $\lambda$.
In a regular statistical model, the learning coefficient is half of the parameter dimension $d/2$.
In LDA, $d/2 = (M+N-1)H/2-N/2$ holds; hence, it linearly increases as the number of topics $H$ does.
On the other hand, the RLCT of LDA $\lambda$ does not.
In addition, $\lambda$ is much smaller than $d/2$.
Fig. \ref{fig:RLCT} shows how the RLCT of LDA $\lambda$ behaves when the number of topics $H$ increases,
with $\lambda$-value in the vertical axis and $H$-value in the horizontal axis.
If $\lambda$ was equal to $d/2$, then it would linearly increase (the square markers dotted plot in Fig. \ref{fig:RLCT}).
However, in fact, $\lambda$ is given by Main Theorem and its curve is obviously non-linear (the circles dotted plot in Fig. \ref{fig:RLCT}). Hence, their values and behaviors are very different.
BIC is based on $d/2$ from the asymptotics of regular statistical models.
In contrast, the foundation of sBIC is singular learning theory; thus, it uses $\lambda$ instead of $d/2$.
That is why sBIC is theoretically recommended for LDA.

We can draw the theoretical learning curve like the solid line in Fig. \ref{fig:LearningCurve},
with $\mathbb{E}[G_n]$-value in the vertical axis and $n$-value in the horizontal axis.
We also namely draw a curve like the dashed line in Fig. \ref{fig:LearningCurve}.
This dashed curve is not only an upper bound of the learning curve of LDA in Bayesian inference
but also a lower bound of that in maximum likelihood or posterior estimation methods.
Let $G_n^{\mathrm{MAP}}$ and $\mu$ be the generalization error and the learning coefficient of LDA in maximum posterior methods, respectively. This is well-defined, i.e. $\mathbb{E}[G_n^{\mathrm{MAP}}] = \mu /n + o(1/n)$ holds.
On the basis of the same prior distribution, Watanabe proved the following inequality \cite{SWatanabeBookMath}:
\begin{equation}
\lambda < d/2 < \mu.
\end{equation}
This means $\mathbb{E}[G_n^{\mathrm{MAP}}] > \mathbb{E}[G_n]  + o(1/n)$
and the leading term of these difference is $(\mu - \lambda)/n > (d-2\lambda)/2n$.
Owing to Main Theorem, we immediately have the exact value of $d-2\lambda$.
Therefore, our result shows at least how much Bayesian inference improves the generalization performance of LDA compared to maximum posterior method.
If the prior distribution is a uniform one, then $\mu$ equals the learning coefficient of LDA in maximum likelihood estimation.
Hence, the above consideration can be applied to maximum likelihood estimation.

One of future works is finding simultaneous resolution of singularities when the prior is a Dirichlet distribution.
A density function of a non-uniform Dirichlet distribution has zero or diverged points; thus, it may affect the learning coefficient.


\appendix
\section{Proof of Main Theorem}
\label{pf-main}

The structure of the proof of Main Theorem is as follows.
First, we summarize terms in $\lVert AB-A_0B_0 \rVert^2$ and consider degeneration of a polynomial ideal.
Second, we resolve the non-negative restriction by variable transformations which are isomorphic maps.
Third, we verify that the problem can be came down to finding the RLCT of reduced rank regression.
Lastly, we calculate the concrete value of the RLCT in each case.

\begin{proof}[Proof of Main Theorem]

As mentioned Sketch of Proof of Main Theorem in Section \ref{sec-main},
we only have to consider the analytic set defined by
$$\{(A,B) \mid \lVert AB-A_0B_0 \rVert^2=0, A \ {\rm and } \ B \ {\rm are \ stochastic \ matrices.}\}$$
to determine the RLCT of LDA $\lambda$ and its multiplicity $m$.

The first part is same as the first half of the proof of Appendix A in our previous research \cite{nhayashi7}.
For the sake of self-containedness, we write down the process of developing the terms in the above paper.
Let $\sim$ be a binomial relation such that the functions $K_1(w)$ and $K_2(w)$ have same RLCT if $K_1(w) \sim K_2(w)$.
Summarizing the terms, we have
\begin{align}\label{Main-1}
\| AB-A_0B_0 \|^2  
&= \sum_{j=1}^{N}
\sum_{i=1}^{M-1}\sum_{k=1}^H a_{ik}b_{kj}-\sum_{k=1}^{H_0} a_{ik}^0b_{kj}^0
+\sum_{j=1}^{N}
\sum_{k=1}^H a_{Mk}b_{kj}-\sum_{k=1}^{H_0} a_{Mk}^0b_{kj}^0. 
\end{align}
Put 
\begin{gather*}
K_{ij}:=\sum_{k=1}^{H}a_{ik}b_{kj} - \sum_{k=1}^{H_0} a^0_{ik}b^0_{kj}, \\
L_j:= \sum_{k=1}^{H} a_{Mk}b_{kj} - \sum_{k=1}^{H_0} a^0_{Mk}b^0_{kj},
\end{gather*}
then we get
$$\|AB-A_0 B_0\|^2 = \sum_{j=1}^N \sum_{i=1}^{M-1}K_{ij}^2 + \sum_{j=1}^N L_j^2.$$
Using $a_{Mk}=1-\sum_{i=1}^{M-1}a_{ik}$, $b_{Hj}=1-\sum_{k=1}^{H-1} b_{kj}$, $a^0_{Mk}=1-\sum_{i=1}^{M-1}a^0_{ik}$, and $b^0_{H_0j}=1-\sum_{k=1}^{H_0-1} b^0_{kj}$, we have
\begin{gather*}
K_{ij} =\sum_{k=1}^{H-1} (a_{ik}-a_{iH})b_{kj} - \sum_{k=1}^{H_0-1} (a^0_{ik}-a^0_{iH_0})b^0_{kj} + (a_{iH}-a^0_{iH_0}), \\
L_j=-\sum_{i=1}^{M-1} \sum_{k=1}^{H-1} (a_{ik}-a_{iH})b_{kj} +\sum_{i=1}^{M-1} \sum_{k=1}^{H_0-1}(a^0_{ik}-a^0_{iH_0}) b^0_{kj}
- \sum_{i=1}^{M-1} (a_{iH}-a^0_{iH_0}),
\end{gather*}
thus
$$L_j^2=\left( \sum_{i=1}^{M-1} K_{ij} \right)^2.$$
Therefore 
\begin{align*}
\| AB-A_0B_0\|^2
&= \sum_{j=1}^N \sum_{i=1}^{M-1} K_{ij}^2 +\sum_{j=1}^N L_j^2 \\
&= \sum_{j=1}^N \sum_{i=1}^{M-1} K_{ij}^2 +\sum_{j=1}^N \left( \sum_{i=1}^{M-1} K_{ij} \right)^2. 
\end{align*}
Since the polynomial $\sum_{i=1}^{M-1} K_{ij} $ is contained in the ideal generated from $(K_{ij})_{i=1,j=1}^{M-1,N}$ , we have
$$\| AB-A_0B_0\|^2 \sim \sum_{j=1}^N \sum_{i=1}^{M-1} K_{ij}^2,$$
i.e.
\begin{align*}
&\quad \| AB-A_0B_0\|^2 \\
&\sim \sum_{j=1}^N \sum_{i=1}^{M-1} \left\{ \sum_{k=1}^{H-1} (a_{ik}-a_{iH})b_{kj} - \sum_{k=1}^{H_0-1} (a^0_{ik}-a^0_{iH_0})b^0_{kj}+ (a_{iH}-a^0_{iH_0}) \right\}^2 \\
&= \sum_{j=1}^N \sum_{i=1}^{M-1} \left[ \sum_{k=1}^{H_0-1} \{(a_{ik}-a_{iH})b_{kj}- (a^0_{ik}-a^0_{iH_0})b^0_{kj}\} + \sum_{k=H_0}^{H-1} (a_{ik}-a_{iH})b_{kj}+ (a_{iH}-a^0_{iH_0}) \right]^2.
\end{align*}
\begin{align}
\label{transform1}
\mbox{Let} \begin{cases}
a_{ik}=a_{ik}-a_{iH}, & k<H \\
c_i = a_{iH}-a^0_{iH_0}, \\
b_{kj}=b_{kj}
\end{cases}
\end{align}
and put $a^0_{ik}=a^0_{ik}-a^0_{iH_0}$. Then we have
\begin{align}
\label{sqerror1}
\| AB-A_0B_0\|^2 
&\sim 
\sum_{j=1}^N \sum_{i=1}^{M-1} \left\{ \sum_{k=1}^{H_0-1} (a_{ik}b_{kj}- a^0_{ik}b^0_{kj}) + \sum_{k=H_0}^{H-1} a_{ik}b_{kj}+ c_i \right\}^2.
\end{align}
This is the end of the common part to Appendix A of \cite{nhayashi7}.
We had derived an upper bound of $\lambda$ by using some inequalities of Frobenius norm and the exact value of $\lambda$ in special cases \cite{nhayashi7}.
However, in this paper, we use changes of variables which resolve non-negative restrictions and find the RLCT in the all cases.

The transformation (\ref{transform1}) resolves the non-negative restrictions of $a_{ik} (k<H)$ and $c_i$ for $i=1,\ldots,M-1$. The changed variables $a_{ik} (k<H)$ and $c_i$ can be negative.
We call the determinant of the Jacobian matrix Jacobian for the sake of simplicity. The Jacobian of the transformation (\ref{transform1}) equals one.
\begin{align}
\label{transform2}
\mbox{Let} \begin{cases}
a_{ik}=a_{ik}, & k<H \\
x_i = c_i + \sum_{k=1}^{H_0-1} (a_{ik}b_{k1}- a^0_{ik}b^0_{k1}) + \sum_{k=H_0}^{H-1} a_{ik}b_{k1}, \\
b_{kj}=b_{kj}.
\end{cases}
\end{align}
It is immediately derived that the Jacobian of this map is equal to one.
About the transform (\ref{transform2}), for $j=2,\ldots,N$, we have
\begin{align}
&\quad \sum_{k=1}^{H_0-1} (a_{ik}b_{kj}- a^0_{ik}b^0_{kj}) + \sum_{k=H_0}^{H-1} a_{ik}b_{kj} + c_i \nonumber \\
&= x_i - \sum_{k=1}^{H_0-1} (a_{ik}b_{k1}- a^0_{ik}b^0_{k1}) - \sum_{k=H_0}^{H-1} a_{ik}b_{k1} + \sum_{k=1}^{H_0-1} (a_{ik}b_{kj}- a^0_{ik}b^0_{kj}) + \sum_{k=H_0}^{H-1} a_{ik}b_{kj}.
\end{align}
Substituting this for $\sum_{k=1}^{H_0-1} (a_{ik}b_{kj}- a^0_{ik}b^0_{kj}) + \sum_{k=H_0}^{H-1} a_{ik}b_{kj} + c_i$ in Eq. (\ref{sqerror1}), we have
\begin{align}
\label{sqerror2}
&\quad \| AB-A_0B_0\|^2 \nonumber \\
&\sim \sum_{i=1}^{M-1} \left\{ \sum_{k=1}^{H_0-1} (a_{ik}b_{k1}- a^0_{ik}b^0_{k1}) + \sum_{k=H_0}^{H-1} a_{ik}b_{k1}+ c_i \right\}^2 \nonumber \\
&\quad + \sum_{j=2}^N \sum_{i=1}^{M-1} \left\{ \sum_{k=1}^{H_0-1} (a_{ik}b_{kj}- a^0_{ik}b^0_{kj}) + \sum_{k=H_0}^{H-1} a_{ik}b_{kj}+ c_i \right\}^2 \nonumber \\
&= \sum_{i=1}^{M-1} x_i^2 + \sum_{j=2}^N \sum_{i=1}^{M-1} \left\{ x_i - \sum_{k=1}^{H_0-1} (a_{ik}b_{k1}- a^0_{ik}b^0_{k1}) - \sum_{k=H_0}^{H-1} a_{ik}b_{k1} \right. \nonumber \\
&\quad \left. + \sum_{k=1}^{H_0-1} (a_{ik}b_{kj}- a^0_{ik}b^0_{kj}) + \sum_{k=H_0}^{H-1} a_{ik}b_{kj} \right\}^2 \nonumber \\
&= \sum_{i=1}^{M-1} x_i^2 + \sum_{j=2}^N \sum_{i=1}^{M-1} \left[ x_i + \sum_{k=1}^{H_0-1} \{a_{ik}(b_{kj}-b_{k1})- a^0_{ik}(b^0_{kj}-b^0_{k1})\} + \sum_{k=H_0}^{H-1} a_{ik}(b_{kj}-b_{k1}) \right]^2.
\end{align}
Put
$$g_{ij} = \sum_{k=1}^{H_0-1} \{a_{ik}(b_{kj}-b_{k1})- a^0_{ik}(b^0_{kj}-b^0_{k1})\} + \sum_{k=H_0}^{H-1} a_{ik}(b_{kj}-b_{k1}).$$
From Eq. (\ref{sqerror2}), we have
\begin{align}
\| AB-A_0B_0\|^2 \sim \sum_{i=1}^{M-1} x_i^2 + \sum_{j=2}^N \sum_{i=1}^{M-1} ( x_i + g_{ij} )^2.
\end{align}
Let $J$ be a polynomial ideal $\langle (x_i)_{i=1}^{M-1}, (g_{ij})_{i=1,j=2}^{M-1,N} \rangle$.
On account of $x_i+g_{ij} \in J$, we have
$$\sum_{i=1}^{M-1} x_i^2 + \sum_{j=2}^N \sum_{i=1}^{M-1} ( x_i + g_{ij} )^2 \sim \sum_{i=1}^{M-1} x_i^2 + \sum_{j=2}^N \sum_{i=1}^{M-1} ( x_i^2 + g_{ij}^2 ),$$
i.e.
\begin{align}
\label{sqerror3}
\| AB-A_0B_0\|^2 &\sim \sum_{j=2}^N \sum_{i=1}^{M-1} ( x_i^2 + g_{ij}^2 ) \nonumber \\
&\sim \sum_{i=1}^{M-1} x_i^2 + \sum_{j=2}^N \sum_{i=1}^{M-1} g_{ij}^2 \nonumber \\
&= \sum_{i=1}^{M-1} x_i^2  + \sum_{j=2}^N \sum_{i=1}^{M-1} \left[ \sum_{k=1}^{H_0-1} \{a_{ik}(b_{kj}-b_{k1})- a^0_{ik}(b^0_{kj}-b^0_{k1})\} + \sum_{k=H_0}^{H-1} a_{ik}(b_{kj}-b_{k1}) \right]^2.
\end{align}
\begin{align}
\label{transform3}
\mbox{Let} \begin{cases}
a_{ik}=a_{ik}, & k<H \\
x_i = x_i, \\
b_{k1}=b_{k1}, \\
b_{kj}=b_{kj}-b_{k1} & j>1.
\end{cases}
\end{align}
For $k=1,\ldots,H-1$ and $j=2,\ldots,N$, non-negative restrictions of $b_{kj}$ can be resolved.
The Jacobian of the transformation (\ref{transform3}) is one.
Apply this map to Eq. (\ref{sqerror3}) and put $b^0_{kj}=b^0_{kj}-b^0_{k1}$. Then, we have
\begin{align}
\label{sqerror4}
\| AB-A_0B_0\|^2 &\sim \sum_{i=1}^{M-1} x_i^2  + \sum_{j=2}^N \sum_{i=1}^{M-1} \left\{ \sum_{k=1}^{H_0-1} (a_{ik}b_{kj}- a^0_{ik}b^0_{kj}) + \sum_{k=H_0}^{H-1} a_{ik}b_{kj} \right\}^2 \nonumber \\
&= \sum_{i=1}^{M-1} x_i^2  + \sum_{j=2}^N \sum_{i=1}^{M-1} \left( \sum_{k=1}^{H-1} a_{ik}b_{kj} - \sum_{k=1}^{H_0-1} a^0_{ik}b^0_{kj} \right)^2.
\end{align}
There are not $b_{k1}$ $(k=1,\ldots,H-1)$ in the right hand side; thus, we can regard the non-negative restrictions of the all variable are resolved after applying the transformation (\ref{transform3}).

Real matrices $U$, $V$, $U_0$, and $V_0$ are denoted by
$U:=(u_{ik})_{i=1, k=1}^{M-1, H-1}$, 
$V:=(v_{kl})_{k=1, l=1}^{H-1, N-1}$,
$U_0:=(u^0_{ik})_{i=1, k=1}^{M-1, H_0-1}$, 
and $V_0:=(v^0_{kl})_{k=1, l=1}^{H_0-1, N-1}$, respectively.
Here, we have
$u_{ik} = a_{ik}$, 
$v_{kl} = v_{k(j-1)} = b_{kj}$, 
$u^0_{ik} = a^0_{ik}$, 
and $v^0_{kl} = v^0_{k(j-1)} = b^0_{kj}$
for $i=1,\ldots,M-1$, $k=1,\ldots,H-1$ and $j=2,\ldots,N$.
Note that these $U_0$ and $V_0$ are same as in Eqs. (\ref{defU0}) and (\ref{defV0})
because of the above transformations for the entries of $A_0$ and $B_0$.
Therefore, $r:=\mathrm{rank}(U_0V_0)$ is also equal to that of Main Theorem.

Now, let us start coming down the problem from LDA to reduced rank regression.
\begin{align}
\label{sqerror-uv}
\| UV-U_0V_0\|^2
&= \sum_{l=1}^{N-1} \sum_{i=1}^{M-1} \left( \sum_{k=1}^{H-1} u_{ik}v_{kl} - \sum_{k=1}^{H_0-1} u^0_{ik}v^0_{kl} \right)^2 \nonumber \\
&= \sum_{j=2}^N \sum_{i=1}^{M-1} \left( \sum_{k=1}^{H-1} a_{ik}b_{kj} - \sum_{k=1}^{H_0-1} a^0_{ik}b^0_{kj} \right)^2
\end{align}
holds; thus, from Eq. (\ref{sqerror4}) and (\ref{sqerror-uv}), we have
\begin{align}
\label{sqerror-last}
\| AB-A_0B_0\|^2 \sim \sum_{i=1}^{M-1} x_i^2  + \| UV-U_0V_0\|^2.
\end{align}
Let $(\lambda_1, m_1)$ and $(\lambda_2, m_2)$ be pairs of the RLCT and its multiplicity of the first and the second term, respectively.
There is no intersection between $\{(x_i)_{i=1}^M\}$ and $\{(U,V)\}$; hence, we have
\begin{gather}
\lambda = \lambda_1 + \lambda_2, \\
m = m_1 + m_2 - 1.
\end{gather}
By simple calculation, $\lambda_1 = (M-1)/2$ and $m_1=1$ hold.
Besides, the entries of the matrices $U$ and $V$ can be real as well as non-negative.
Thus, $\lambda_2$ is the RLCT of non-restricted MF, i.e. that of reduced rank regression \cite{Aoyagi1}.
The same is true for the multiplicity $m_2$.
Therefore, we obtain
\begin{gather}
\label{thm-RLCT-addform}
\lambda = \frac{M-1}{2} + \lambda_{\mathrm{MF}}(M-1,N-1,H-1,r), \\
m = m_{\mathrm{MF}}(M-1,N-1,H-1,r),
\end{gather}
where $r=\mathrm{rank}(U_0V_0)$.

Finally, we concretely calculate $\lambda$ and $m$.
According to \cite{Aoyagi1}, the RLCT and its multiplicity of MF are as follows.\\
(1) If $N+r+1\leqq M+H$ and $M+r+1 \leqq N+H$ and $H+r+1 \leqq M+N$ and $M+N+H+r+1$ is even ($M+N+H+r$ is odd), then
\begin{gather*}
\lambda_{\mathrm{MF}}(M-1,N-1,H-1,r) = \frac{1}{8}\{ 2(H+r-1)(M+N-2)-(M-N)^2-(H+r-1)^2 \}, \\
m_{\mathrm{MF}}(M-1,N-1,H-1,r) = 1.
\end{gather*}
(2) Else if $N+r+1\leqq M+H$ and $M+r+1 \leqq N+H$ and $H+r+1 \leqq M+N$ and $M+N+H+r+1$ is odd ($M+N+H+r$ is even), then
\begin{gather*}
\lambda_{\mathrm{MF}}(M-1,N-1,H-1,r) = \frac{1}{8}\{ 2(H+r-1)(M+N-2)-(M-N)^2-(H+r-1)^2 +1 \}, \\
m_{\mathrm{MF}}(M-1,N-1,H-1,r) = 2.
\end{gather*}
(3) Else if $M+H<N+r+1$, then
\begin{gather*}
\lambda_{\mathrm{MF}}(M-1,N-1,H-1,r) = \frac{1}{2}\{(M-1)(H-1)+(N-1)r-(H-1)r\}, \\
m_{\mathrm{MF}}(M-1,N-1,H-1,r) = 1.
\end{gather*}
(4) Else if $N+H<M+r+1$, then
\begin{gather*}
\lambda_{\mathrm{MF}}(M-1,N-1,H-1,r) = \frac{1}{2}\{(N-1)(H-1)+(M-1)r-(H-1)r\}, \\
m_{\mathrm{MF}}(M-1,N-1,H-1,r) = 1.
\end{gather*}
(5) Else (i.e. $M+N<H+r+1$), then
\begin{gather*}
\lambda_{\mathrm{MF}}(M-1,N-1,H-1,r) = \frac{1}{2}(M-1)(N-1), \\
m_{\mathrm{MF}}(M-1,N-1,H-1,r) = 1.
\end{gather*}
Since the multiplicity is clear, we find the RLCT.
We develop the terms in each case by using Eq. (\ref{thm-RLCT-addform}). \\
In the case (1), we have
\begin{align}
\lambda &= \frac{M-1}{2} + \frac{1}{8}\{ 2(H+r-1)(M+N-2)-(M-N)^2-(H+r-1)^2 \} \\
&= \frac{M-1}{2} + \frac{1}{8}\{ 2(H+r)(M+N)-2(M+N)-4(H+r)+4 \nonumber \\
&\quad -(M-N)^2-(H+r)^2 +2(H+r) -1 \} \\
&= \frac{1}{8}\{ 4M-4 + 2(H+r)(M+N)-(M-N)^2-(H+r)^2 -2(M+N)-2(H+r)+3 \} \\
&= \frac{1}{8}\{ 2(H+r)(M+N)-(M-N)^2-(H+r)^2 +2(M+N)-2(H+r)-1 -4N \} \\
&= \frac{1}{8}\{ 2(H+r+1)(M+N)-(M-N)^2-(H+r+1)^2 \} - \frac{N}{2}.
\end{align}
In the case (2), by the same way as the case (1), we have
\begin{align}
\lambda = \frac{1}{8}\{ 2(H+r+1)(M+N)-(M-N)^2-(H+r+1)^2+1 \} - \frac{N}{2}.
\end{align}
In the case (3), we have
\begin{align}
\lambda &= \frac{M-1}{2} + \frac{1}{2}\{(M-1)(H-1)+(N-1)r-(H-1)r\} \\
&= \frac{M-1}{2} + \frac{1}{2}\{MH-(M+H)+1+Nr-r -Hr +r\} \\
&= \frac{1}{2}(MH-M-H+1+Nr-Hr + M-1 +N-N) \\
&= \frac{1}{2}\{ MH+1+N(r+1)-H(r+1) -N \} \\
&= \frac{1}{2}\{ MH+1+N(r+1)-H(r+1) \} - \frac{N}{2}.
\end{align}
In the case (4), by the same way as the case (3), we have
\begin{align}
\lambda = \frac{1}{2}\{ NH+1+M(r+1)-H(r+1) \} - \frac{N}{2}.
\end{align}
In the case (5), we have
\begin{align}
\lambda &= \frac{M-1}{2} + \frac{1}{2}(M-1)(N-1) \\
&= \frac{1}{2}(M-1)N \\
&= \frac{1}{2}MN - \frac{N}{2}.
\end{align}
From the above, Main Theorem is proved. Comparing the RLCT of MF \cite{Aoyagi1}, we also obtain
$$\lambda = \lambda_{\mathrm{MF}}(M,N,H,H_0)-\frac{N}{2}.$$
\end{proof}

\section*{Acknowledgments}

The author would like to express his appreciation to the editor and the reviewers for pointing out ways to improve this paper.



\bibliography{bibs-full}

\begin{thebibliography}{10}

\bibitem{Akaike1974AIC}
Hirotugu Akaike.
\newblock A new look at the statistical model identification.
\newblock {\em IEEE transactions on automatic control}, 19(6):716--723, 1974.

\bibitem{Aoyagi2}
Miki Aoyagi.
\newblock Stochastic complexity and generalization error of a restricted
  boltzmann machine in bayesian estimation.
\newblock {\em Journal of Machine Learning Research}, 11(Apr):1243--1272, 2010.

\bibitem{Aoyagi3}
Miki Aoyagi.
\newblock Learning coefficient in bayesian estimation of restricted boltzmann
  machine.
\newblock {\em Journal of Algebraic Statistics}, 4(1):30--57, 2013.

\bibitem{Aoyagi1}
Miki Aoyagi and Sumio Watanabe.
\newblock Stochastic complexities of reduced rank regression in bayesian
  estimation.
\newblock {\em Neural Networks}, 18(7):924--933, 2005.

\bibitem{Atiyah1970resolution}
Michael~Francis Atiyah.
\newblock Resolution of singularities and division of distributions.
\newblock {\em Communications on pure and applied mathematics}, 23(2):145--150,
  1970.

\bibitem{Bernstein1972}
Joseph Bernstein.
\newblock The analytic continuation of generalized functions with respect to a
  parameter.
\newblock {\em Funktsional'nyi Analiz i ego Prilozheniya}, 6(4):26--40, 1972.

\bibitem{Bezanson2017julia}
Jeff Bezanson, Alan Edelman, Stefan Karpinski, and Viral~B Shah.
\newblock Julia: A fresh approach to numerical computing.
\newblock {\em SIAM review}, 59(1):65--98, 2017.

\bibitem{David2003LDA}
David~M Blei, Andrew~Y Ng, and Michael~I Jordan.
\newblock Latent dirichlet allocation.
\newblock {\em Journal of machine Learning research}, 3(Jan):993--1022, 2003.

\bibitem{Drton2017forest}
Mathias Drton, Shaowei Lin, Luca Weihs, Piotr Zwiernik, et~al.
\newblock Marginal likelihood and model selection for gaussian latent tree and
  forest models.
\newblock {\em Bernoulli}, 23(2):1202--1232, 2017.

\bibitem{Drton}
Mathias Drton and Martyn Plummer.
\newblock A bayesian information criterion for singular models.
\newblock {\em Journal of the Royal Statistical Society Series B}, 79:323--380,
  2017.
\newblock with discussion.

\bibitem{Gildea1999topic}
Daniel Gildea and Thomas Hofmann.
\newblock Topic-based language models using em.
\newblock In {\em Sixth European Conference on Speech Communication and
  Technology}, 1999.

\bibitem{Griffiths2004LDAGS}
Thomas~L Griffiths and Mark Steyvers.
\newblock Finding scientific topics.
\newblock {\em Proceedings of the National academy of Sciences}, 101(suppl
  1):5228--5235, 2004.

\bibitem{nhayashi8}
Naoki Hayashi.
\newblock Variational approximation error in non-negative matrix factorization.
\newblock {\em Neural Networks}, 126:65--75, 2020.

\bibitem{nhayashi5}
Naoki Hayashi and Sumio Watanabe.
\newblock Tighter upper bound of real log canonical threshold of non-negative
  matrix factorization and its application to bayesian inference.
\newblock In {\em IEEE Symposium Series on Computational Intelligence (IEEE
  SSCI)}, pages 718--725, 11 2017.

\bibitem{nhayashi2}
Naoki Hayashi and Sumio Watanabe.
\newblock Upper bound of bayesian generalization error in non-negative matrix
  factorization.
\newblock {\em Neurocomputing}, 266C(29 November):21--28, 2017.

\bibitem{nhayashi7}
Naoki Hayashi and Sumio Watanabe.
\newblock Asymptotic bayesian generalization error in latent dirichlet
  allocation and stochastic matrix factorization.
\newblock {\em SN Computer Science}, 1(2):1--22, 2020.

\bibitem{Hironaka}
Heisuke Hironaka.
\newblock Resolution of singularities of an algbraic variety over a field of
  characteristic zero.
\newblock {\em Annals of Mathematics}, 79:109--326, 1964.

\bibitem{Imai2019estimating}
Toru Imai.
\newblock Estimating real log canonical thresholds.
\newblock {\em arXiv preprint arXiv:1906.01341}, 2019.

\bibitem{Li2005LDA4CV}
Fei-Fei Li and Pietro Perona.
\newblock A bayesian hierarchical model for learning natural scene categories.
\newblock In {\em Proceedings of the 2005 IEEE Computer Society Conference on
  Computer Vision and Pattern Recognition (CVPR'05) - Volume 02}, pages
  524--531, Washington, DC, USA, 2005. IEEE Computer Society.

\bibitem{Matsuda1-e}
Ken Matsuda and Sumio Watanabe.
\newblock Weighted blowup and its application to a mixture of multinomial
  distributions.
\newblock {\em IEICE Transactions}, J86-A(3):278--287, 2003.
\newblock in Japanese.

\bibitem{Nagata2008asymptotic}
Kenji Nagata and Sumio Watanabe.
\newblock Asymptotic behavior of exchange ratio in exchange monte carlo method.
\newblock {\em Neural Networks}, 21(7):980--988, 2008.

\bibitem{Rusakov2005asymptotic}
Dmitry Rusakov and Dan Geiger.
\newblock Asymptotic model selection for naive bayesian networks.
\newblock {\em Journal of Machine Learning Research}, 6(Jan):1--35, 2005.

\bibitem{SatoK2019PMM}
Kenichiro Sato and Sumio Watanabe.
\newblock Bayesian generalization error of poisson mixture and simplex
  vandermonde matrix type singularity.
\newblock {\em arXiv preprint arXiv:1912.13289}, 2019.

\bibitem{Sato1974zeta}
Mikio Sato and Takuro Shintani.
\newblock On zeta functions associated with prehomogeneous vector spaces.
\newblock {\em Annals of Mathematics}, pages 131--170, 1974.

\bibitem{Schwarz1978BIC}
Gideon Schwarz.
\newblock Estimating the dimension of a model.
\newblock {\em The annals of statistics}, 6(2):461--464, 1978.

\bibitem{Tirunillai2014LDA4MR}
Seshadri Tirunillai and Gerard~J Tellis.
\newblock Mining marketing meaning from online chatter: Strategic brand
  analysis of big data using latent dirichlet allocation.
\newblock {\em Journal of Marketing Research}, 51(4):463--479, 2014.

\bibitem{Watanabe1}
Sumio Watanabe.
\newblock Algebraic analysis for non-regular learning machines.
\newblock {\em Advances in Neural Information Processing Systems}, 12:356--362,
  2000.
\newblock Denver, USA.

\bibitem{Watanabe2}
Sumio Watanabe.
\newblock Algebraic geometrical methods for hierarchical learning machines.
\newblock {\em Neural Networks}, 13(4):1049--1060, 2001.

\bibitem{SWatanabeBookE}
Sumio Watanabe.
\newblock {\em Algebraix Geometry and Statistical Learning Theory}.
\newblock Cambridge University Press, 2009.

\bibitem{WatanabeAIC}
Sumio Watanabe.
\newblock Asymptotic equivalence of bayes cross validation and widely
  applicable information criterion in singular learning theory.
\newblock {\em Journal of Machine Learning Research}, 11(Dec):3571--3594, 2010.

\bibitem{WatanabeBIC}
Sumio Watanabe.
\newblock A widely applicable bayesian information criterion.
\newblock {\em Journal of Machine Learning Research}, 14(Mar):867--897, 2013.

\bibitem{SWatanabeBookMath}
Sumio Watanabe.
\newblock {\em Mathematical theory of Bayesian statistics}.
\newblock CRC Press, 2018.

\bibitem{Yamazaki1}
Keisuke Yamazaki and Sumio Watanabe.
\newblock Singularities in mixture models and upper bounds of stochastic
  complexity.
\newblock {\em Neural Networks}, 16(7):1029--1038, 2003.

\bibitem{Yamazaki3}
Keisuke Yamazaki and Sumio Watanabe.
\newblock Stochastic complexity of bayesian networks.
\newblock In {\em Uncertainty in Artificial Intelligence (UAI'03)}, 2003.

\bibitem{Yamazaki2}
Keisuke Yamazaki and Sumio Watanabe.
\newblock Algebraic geometry and stochastic complexity of hidden markov models.
\newblock {\em Neurocomputing}, 69:62--84, 2005.
\newblock issue 1-3.

\bibitem{Yamazaki4}
Keisuke Yamazaki and Sumio Watanabe.
\newblock Singularities in complete bipartite graph-type boltzmann machines and
  upper bounds of stochastic complexities.
\newblock {\em IEEE Transactions on Neural Networks}, 16:312--324, 2005.
\newblock issue 2.

\bibitem{Yoshida2018LDA4Geo}
Kenta Yoshida, Tatsu Kuwatani, Takao Hirajima, Hikaru Iwamori, and Shotaro
  Akaho.
\newblock Progressive evolution of whole-rock composition during metamorphism
  revealed by multivariate statistical analyses.
\newblock {\em Journal of Metamorphic Geology}, 36(1):41--54, 2018.

\bibitem{Zwiernik2011asymptotic}
Piotr Zwiernik.
\newblock An asymptotic behaviour of the marginal likelihood for general markov
  models.
\newblock {\em Journal of Machine Learning Research}, 12(Nov):3283--3310, 2011.

\end{thebibliography}
\bibliographystyle{plain}






\end{document}